\documentclass[conference]{IEEEtran}
\IEEEoverridecommandlockouts
\usepackage{cite}
\usepackage{amsmath, amssymb, amsfonts}
\usepackage{algorithmic}
\usepackage{graphicx}
\usepackage{hyperref}
\usepackage{textcomp}
\usepackage{xcolor}
\usepackage{url}
\usepackage{makecell}

\def\BibTeX{{\rm B\kern-.05em{\sc i\kern-.025em b}\kern-.08em T\kern-.1667em\lower.7ex\hbox{E}\kern-.125emX}}
\begin{document}
    \title{A Cost-Benefit Analysis of On-Premise Large Language Model Deployment: Breaking Even with Commercial LLM Services}

    \author{\IEEEauthorblockN{Guanzhong Pan}
\IEEEauthorblockA{\textit{Carnegie Mellon University}\\
arthurp@andrew.cmu.edu}
\and \IEEEauthorblockN{Vishal Chodnekar}
\IEEEauthorblockA{\textit{Unaffiliated}\\
vishal.chodnekar@gmail.com}
\and \IEEEauthorblockN{Abinas Roy}
\IEEEauthorblockA{\textit{Unaffiliated}\\
Abinasroy@gmail.com}
\and \IEEEauthorblockN{Haibo Wang}
\IEEEauthorblockA{\textit{Carnegie Mellon University}\\
haibowan@alumni.cmu.edu}
    }

    \maketitle

    \begin{abstract}
        Large language models (LLMs) are becoming increasingly widespread. Organizations that want to use AI for productivity now face an important decision. They can subscribe to commercial LLM services or deploy models on their own infrastructure. Cloud services from providers such as OpenAI, Anthropic, and Google are attractive because they provide easy access to state-of-the-art models and are easy to scale. However, concerns about data privacy, the difficulty of switching service providers, and long-term operating costs have driven interest in local deployment of open-source models. This paper presents a cost-benefit analysis framework to help organizations determine when on-premise LLM deployment becomes economically viable compared to commercial subscription services. We consider the hardware requirements, operational expenses, and performance benchmarks of the latest open-source models, including Qwen, Llama, Magistral, and etc. Then we compare the total cost of deploying these models locally with the major cloud providers subscription fee. Our findings provide an estimated breakeven point based on usage levels and performance needs. These results give organizations a practical framework for planning their LLM strategies.
    \end{abstract}


    \section{Introduction}
    The rapid development of Large Language Models(LLMs) has driven organizations to apply them in user-facing services for more adaptive user-facing services~\cite{shareef2024, zhao2023, minaee2024}. As adoption grows, organizations face a critical strategic choice: whether to rely on commercial cloud-based services subscription or to invest in their own on-premise deployment infrastructure~\cite{zhang2025, huang2024}. Providers such as OpenAI, Anthropic, and Google offer easy API access to state-of-the-art and agentic models~\cite{luqman2025agentic}, but costs scale quickly with usage~\cite{chen2023}. These solutions also raise concerns over compliance, data protection, and the challenge of transferring service providers~\cite{chen2025, dennstadt2025}. For example, Desai et al.~\cite{desai2024} note that privacy issues hinder LLM adoption in finance, where trust and regulation are critical.

    At the same time, organizations now have the option to deploy open-source models, including LLaMA~\cite{llama}, Mistral~\cite{mistral}, and Qwen~\cite{qwen}. Recent advances in GPU hardware (NVIDIA H100\cite{nvidiaH100}, AMD MI300X\cite{amdMI300XProduct}) and inference optimization frameworks (vLLM\cite{kwon2023}, NVIDIA TensorRT-LLM\cite{tensorrt2024}, DeepSpeed\cite{yazdani2022}) also contributed to making local deployment more feasible. As a result, it is worth for enterprises to reconsider whether building their own AI infrastructure may be a more economically viable option.

    Despite rising interest in hybrid deployment, few systematic and quantitative comparisons of the two approaches exist. Organizations need a framework to determine when local deployment outweighs commercial services in cost-effectiveness.

    This paper provides a comprehensive cost-benefit analysis framework for on-premise open-source LLM deployment. Our main contributions are:
    \begin{enumerate}
        \item A survey of commercial LLM pricing models and open-source alternatives suitable for local deployment.
        \item Mathematical models for total cost of ownership (TCO) analysis comparing local open-source LLM deployment and commercial API usage.
        \item A playground\footnote{Available at: \url{https://v0-ai-cost-calculator.vercel.app/}} where enterprise users can apply the cost-benefit framework to explore hardware/API trade-offs.

    \end{enumerate}
    
    Our analysis reveals that on-premise deployment are economically viable, with break-even periods typically within a few months for small models, 2 years for medium models and 5 years for larger models, making it viable primarily for organizations with extreme high-volume processing requirements ($\ge$50M tokens/month) or strict data residency mandates.

    \section{Background and Related Work}
    
    \subsection{LLM Deployment and Cost Trade-offs}

    Modern LLM deployment spans three paradigms: \textit{cloud}, \textit{on-premise}, and \textit{hybrid}. Cloud services provide immediate access to state-of-the-art models but raise issues of recurring cost, data security, and sovereignty~\cite{zhao2023survey,park2025survey}. On-premise deployments offer full control and compliance for sensitive domains such as healthcare, finance, and law~\cite{wu2025mia}, yet require significant upfront investment. Hybrid solutions balance these trade-offs by running critical workloads locally while offloading scalable or latency-sensitive tasks to the cloud~\cite{hao2024hybrid}.
    
    Prior cost analyses of LLM deployment focus on four directions: (1) reducing API expenditure through pricing strategies and model cascades~\cite{chen2023,overthink_cost}, (2) minimizing inference cost per token via quantization, batching, and speculative decoding~\cite{kwon2023,deepspeed_inf,flexgen,leviathan_specdec}, (3) improving system efficiency through serverless or multi-tenant provisioning~\cite{serverlessllm,azure_ptu_doc,slora_mlsys,medusa_serverless}, and (4) modeling TCO considering hardware amortization, precision formats, and energy efficiency~\cite{inference_econ,tco_fp8,energy_considerations}. 
    
    Despite these advances, few frameworks jointly analyze the economics of cloud and local deployment. Existing studies typically isolate API or infrastructure costs and leave a gap in unified evaluation. This work addresses this gap by integrating pricing, performance, and governance into a single cost-benefit framework for LLM deployment.

    \subsection{Open-Source Model Ecosystem}
    
    The open-source LLM ecosystem has accelerated, with many models achieving performance as well as commercial models. Meta’s LLaMA-3 family (8B–405B parameters) matches or surpasses models like Claude and Gemini in reasoning and knowledge tests~\cite{touvron2023llama}. Similarly, Alibaba’s open-source Qwen-3 sets new benchmarks in mathematics, coding, and multilingual understanding~\cite{bai2023qwen}. The rapid spread of such models allows organizations to run advanced LLMs in-house for greater cost control, feature customization, and data security. As a result, many formerly reliant on commercial APIs are now considering open-source alternatives.

    \section{Survey Methodology and Model Selection}
    
    \subsection{Performance Evaluation Framework}
    We developed a performance evaluation framework to ensure our cost-benefit analysis reflects deployment conditions. 

    \textbf{Accuracy Metrics.} 
    While accuracy is essential, enterprises require evidence of task-specific performance. We evaluate models across diverse benchmarks encompassing reasoning, mathematics, coding, and multi-domain understanding~\cite{desai2024emerging,rein2023,hendrycks2021,wang2024,jain2024}. This captures analytical, computational, and applied challenges representative of enterprise workloads. 
    
    We then compare open-weight models (e.g., LLaMA, Qwen-3) with commercial models (e.g., ChatGPT, Claude) to quantify performance gaps and assess the feasibility of replacing APIs with open-source alternatives. Results are presented in Table~\ref{tab:benchmark} (adapted from~\cite{artificialanalysis2025}) and summarized in Table~\ref{tab:breakeven-summary}. This evaluation in realistic tasks ensures organization's model selection aligns with their needs and budget.
    
    \subsection{Model Selection Criteria}
    Integrating accuracy and operational factors, we define four criteria for model inclusion:

    \begin{enumerate}
        \item \textbf{Performance Parity:} Benchmark scores within 20\% of top commercial models, reflecting enterprise norms where small accuracy gaps are offset by cost, security, and integration benefits.
        \item \textbf{Deployment Feasibility:} Hardware requirements suitable for typical enterprise infrastructure.
        \item \textbf{License Compatibility:} Open-weight models under permissive licenses enabling commercial use.
        \item \textbf{Community Support:} Active development communities ensuring continuous improvement and stability.
    \end{enumerate}

    \begin{table*}[htbp]
    \caption{Performance Benchmarks and API Pricing for Leading LLMs (data from Artificial Analysis~\cite{artificialanalysis2025})}
    \label{tab:benchmark}
    \centering
    \small
    \begin{tabular}{|c|c|c|c|c|c|c|}
    \hline
    \textbf{\makecell{Model}} &
    \textbf{\makecell{Total\\Params}} &
    \textbf{GPQA} &
    \textbf{MATH-500} &
    \textbf{LiveCodeBench} &
    \textbf{MMLU-Pro} &
    \textbf{\makecell{API Cost\\(In/Out, USD / 1M tokens)}} \\
    \hline
    \multicolumn{7}{|c|}{\textit{Large Open Models}} \\
    \hline
    Kimi-K2~\cite{kimi2025, kimi_hf2025} & 1T & 76.6\% & 97.1\% & 55.6\% & 82.4\% & -- \\
    GLM-4.5~\cite{glm2025, glm_hf2025} & 355B & 78.2\% & 97.9\% & 73.8\% & 83.5\% & -- \\
    Qwen3-235B~\cite{qwen3_2025, qwen_hf2025} & 235B & 79.0\% & 98.4\% & 78.8\% & 84.3\% & -- \\
    \hline
    \multicolumn{7}{|c|}{\textit{Medium Open Models}} \\
    \hline
    gpt-oss-120B~\cite{gptoss2025, gptoss_hf2025} & 120B & 78.2\% & -- & 63.9\% & 80.8\% & -- \\
    GLM-4.5-Air~\cite{glm2025, glm45air_hf2025} & 106B & 73.3\% & 96.5\% & 68.4\% & 81.5\% & -- \\
    Llama-3.3-70B~\cite{llama3_2024, llama33_hf2024} & 70B & 49.8\% & 77.3\% & 28.8\% & 71.3\% & -- \\
    \hline
    \multicolumn{7}{|c|}{\textit{Small Open Models}} \\
    \hline
    EXAONE 4.0 32B~\cite{exaone2025, exaone_hf2025} & 32B & 73.9\% & 97.7\% & 74.7\% & 81.8\% & -- \\
    Qwen3-30B~\cite{qwen3_2025, qwen30b_hf2025} & 30B & 70.7\% & 97.6\% & 70.7\% & 80.5\% & -- \\
    Magistral Small~\cite{magistral2025, magistral_small_hf2025} & 24B & 64.1\% & 96.3\% & 51.4\% & 74.6\% & -- \\
    \hline
    \multicolumn{7}{|c|}{\textit{Commercial Reference}} \\
    \hline
    GPT-5 (by OpenAI~\cite{openai-pricing}) & -- & 85.4\% & \textbf{99.4\%} & 66.8\% & \textbf{87.1\%} & \$1.25 / \$10.00 \\
    Claude-4 Opus (by Anthropic~\cite{anthropic-pricing}) & -- & 70.1\% & 94.1\% & 54.2\% & 86.0\% & \$15.00 / \$75.00 \\
    Claude-4 Sonnet (by Anthropic~\cite{anthropic-pricing}) & -- & 68.3\% & 93.4\% & 44.9\% & 83.7\% & \$3.00 / \$15.00 \\
    Grok-4 (by xAI~\cite{xai_pricing}) & -- & \textbf{87.7\%} & 99.0\% & \textbf{81.9\%} & 86.6\% & \$3.00 / \$15.00 \\
    Gemini 2.5 Pro (by Google~\cite{gemini_pricing}) & -- & 84.4\% & 96.7\% & 80.1\% & 86.2\% & \$1.25 / \$10.00 \\
    \hline

    \end{tabular}
    \end{table*}
    
    Table~\ref{tab:benchmark} presents our selected models and their benchmark performance compared to commercial alternatives.

    \section{Commercial LLM Pricing Models}
    The API subscription model charges per processed token (input and output), making it suitable for integrating LLMs into custom applications. Costs vary by usage, batching, and model choice. Detailed breakdowns appear in Table~\ref{tab:benchmark}.
    
    \section{On-Premise Deployment: A Cost Breakdown}
    On-premise deployment means running LLMs entirely using an organization’s own data centers or specially designed hardware. This approach does not require any external cloud providers, which provides full control of privacy. We decompose the cost structure into these 3 aspects:
    \begin{itemize}
        \item \textbf{Capital Expenditures (CapEx)}: Hardware (GPUs, servers,
            storage), initial setup, networking.\cite{kim2025tco}

        \item \textbf{Operational Expenditures (OpEx)}: Electricity, cooling, maintenance,
            personnel, software licensing.\cite{kim2025tco}

        \item \textbf{Scaling Costs}: Additional hardware and operational costs as
            user base or workload grows.
    \end{itemize}

    \subsection{Capital Expenditures: The Hardware}
    The largest upfront cost is compute infrastructure. GPU selection is crucial:
    \begin{itemize}
        \item \textbf{Data Center Grade (e.g., NVIDIA A100\cite{nvidia_a100
        })}: Great performance for large-scale, multi-user deployments. Enables running the largest models and highest throughput.

        \item \textbf{Prosumer/Workstation Grade (e.g., RTX 5090\cite{nvidia_rtx5090})}: Suitable
            for small teams or research, lower cost but limited scalability.
    \end{itemize}

    \begin{table}[htbp]
    \caption{Comparison of A100-80GB and RTX 5090-32GB GPUs}
    \label{tab:gpu-comparison}
    \centering
    \begin{tabular}{|l|l|r|r|}
        \hline
        \textbf{GPU} & \textbf{Memory} & \textbf{Power} & \textbf{Price (USD)} \\
        \hline
        NVIDIA RTX 5090-32GB\cite{nvidia_rtx5090} & 32 GB & 575 W & \$2,000 \\
        \hline
        NVIDIA A100-80GB\cite{nvidia_a100
        } & 80 GB & 400 W & \$15,000 \\
        \hline
    \end{tabular}
\end{table}

    Table~\ref{tab:gpu-comparison} lists the memory, power, and approximate price for NVIDIA 5090-32GB and A100-80GB GPUs. Table~\ref{tab:hardware} lists the hardware resources needed to set up the service.

\begin{table*}[htbp]
    \caption{Hardware and Capacity Summary for Deployable Open-Source Models (FP8/W8A16 on A100\cite{nvidia_a100}/RTX 5090\cite{nvidia_rtx5090}, Electricity at \$0.15/kWh)}
    \label{tab:hardware}
    \centering
    \small
    \begin{tabular}{|c|c|c|c|c|c|c|}
    \hline
    \textbf{\makecell{Model}} &
    \textbf{\makecell{MoE}} &
    \textbf{\makecell{VRAM\\(FP8)}} &
    \textbf{\makecell{Hardware\\Deployment}} &
    \textbf{\makecell{Throughput\\(tok/sec)}} &
    \textbf{\makecell{Hardware\\Cost}} &
    \textbf{\makecell{Token\\Capacity/Month}} \\
    \hline
    \multicolumn{7}{|c|}{\textit{Large Open Models}} \\
    \hline
    Kimi-K2~\cite{kimi_hf2025} & Yes & 1000 GB & 16$\times$ A100-80GB & 800 & \$240k & 506.9M \\
    \hline
    GLM-4.5~\cite{glm_hf2025} & Yes & 355 GB & 6$\times$ A100-80GB & 400 & \$90k & 253.4M \\
    \hline
    Qwen3-235B~\cite{qwen_hf2025} & Yes & 235 GB & 4$\times$ A100-80GB & 400 & \$60k & 253.4M \\
    \hline
    \multicolumn{7}{|c|}{\textit{Medium Open Models}} \\
    \hline
    gpt-oss-120B~\cite{gptoss_hf2025} & Yes & 120 GB & 2$\times$ A100-80GB & 220 & \$30k & 139.4M \\
    \hline
    GLM-4.5-Air~\cite{glm45air_hf2025} & Yes & 106 GB & 2$\times$ A100-80GB & 200 & \$30k & 126.7M \\
    \hline
    Llama-3.3-70B~\cite{llama33_hf2024} & No & 70 GB & 1$\times$ A100-80GB & 190 & \$15k & 120.4M \\
    \hline
    \multicolumn{7}{|c|}{\textit{Small Open Models}} \\
    \hline
    EXAONE 4.0 32B~\cite{exaone_hf2025} & No & 32 GB & 1$\times$ RTX 5090 & 200 & \$2k & 126.7M \\
    \hline
    Qwen3-30B~\cite{qwen30b_hf2025} & No & 30 GB & 1$\times$ RTX 5090 & 180 & \$2k & 114.0M \\
    \hline
    Magistral Small~\cite{magistral_small_hf2025} & No & 24 GB & 1$\times$ RTX 5090 & 150 & \$2k & 95.0M \\
    \hline
    \end{tabular}
\end{table*}

    \section{Cost Model and Break-Even Analysis}
    We developed a quantitative model to assess the cost of matching the throughput of leading commercial LLMs. Our model uses available benchmark, hardware, and pricing data.

    \subsection{Cost-Performance Tradeoffs of Open-Source LLM Deployment}

Tables~\ref{tab:benchmark} and~\ref{tab:breakeven-summary} show that open-weight models can deliver competitive performance. Although large open models such as Kimi-K2, GLM-4.5, and Qwen3-235B require GPU clusters costing over \$200k, their accuracy on enterprise benchmarks places them close to leading closed models. This suggests that, technically, open alternatives can approach commercial state-of-the-art, albeit with higher operational complexity.

Notably, the gap between “large” and “medium” deployments is smaller than expected. Medium-scale models like \textit{gpt-oss-120B}, \textit{GLM-4.5-Air}, and \textit{Llama-3.3-70B} run efficiently on two A100-80GB GPUs (\$30k), with less than 10\% accuracy loss. They thus offer substantially lower ownership costs while maintaining strong performance across reasoning, coding, and domain-specific tasks.

Small models such as \textit{EXAONE 4.0 32B}, \textit{Qwen3-30B}, and \textit{Magistral Small} show that sub-30B deployments are feasible on a single consumer-grade RTX 5090 (\$2k). With performance comparable to medium models, they suit small and mid-sized enterprises prioritizing cost efficiency and local control. Although they trail the highest benchmarks, the practical performance gap between 30B- and 70B-class models remains modest, indicating that smaller deployments can meet a wide range of enterprise needs.

While commercial APIs still hold a slight edge in peak accuracy and efficiency, open-weight models now provide viable, cost-effective alternatives. The performance gaps across large, medium, and small open deployments are far narrower than their order-of-magnitude hardware cost differences. For many organizations, deploying a medium or even small open model locally offers a sustainable break-even option balancing capability, cost, and autonomy from external providers.
\subsection{Cost Model}

The one-time infrastructure cost is defined as:
\begin{equation}
    C_{\text{hardware}} = N_{\text{GPU}} \cdot C_{\text{GPU}}
    \label{eq:hardware_cost}
\end{equation}

Assuming business operation of 8 hours/day and 20 days/month, the monthly electricity cost is:
\begin{equation}
    C_{\text{electricity}} = N_{\text{GPU}} \cdot P_{\text{GPU}} \cdot H_{\text{operation}} \cdot R_{\text{electricity}}
    \label{eq:electricity_cost}
\end{equation}

The total local deployment cost becomes:
\begin{equation}
    C_{\text{local}}(t) = C_{\text{hardware}} + C_{\text{electricity}} \cdot t
    \label{eq:local_cost}
\end{equation}

\noindent where $t$ is the number of months after deployment, $C_{\text{GPU}}$ and $N_{\text{GPU}}$ denote per-unit GPU cost and quantity, $P_{\text{GPU}}$ power consumption, $R_{\text{electricity}}$ the electricity rate, and $H_{\text{operation}}$ the monthly operating hours.

\subsection{Commercial API Throughput Analysis}

For a fair comparison, API costs are normalized by the token generation capacity $Q_{\text{capacity}}$ achievable on local hardware:
\begin{equation}
    Q_{\text{capacity}} = T_{\text{throughput}} \cdot H_{\text{operation}} \cdot 3600
    \label{eq:capacity}
\end{equation}

The equivalent API cost for producing the same token volume per month is:
\begin{equation}
    C_{\text{API}}(Q_{\text{capacity}}) = 
    \frac{Q_{\text{capacity}}}{3} \cdot \frac{C_{\text{input}}}{1M}
    + \frac{2Q_{\text{capacity}}}{3} \cdot \frac{C_{\text{output}}}{1M}
\end{equation}

and scales linearly with time:
\begin{equation}
    C_{\text{API}}(t) = C_{\text{API}}(Q_{\text{capacity}}) \cdot t
\end{equation}

Here $C_{\text{input}}$ and $C_{\text{output}}$ are API prices per million input/output tokens, using a 2:1 ratio typical of real workloads. 

\subsection{Break-Even Analysis}

Local and API costs are modeled as time-dependent functions:
\begin{equation}
    C_{\text{local}}(t) = C_{\text{API}}(t)
    \label{equation:break-even}
\end{equation}

Solving yields the break-even time $t^{*}$, where cumulative local and API costs are equal. For $t > t^{*}$, local deployment becomes more economical, particularly under sustained, high-throughput usage. The next section reports $t^{*}$ values across open-source and commercial models.

    \subsection{Comprehensive Break-Even Analysis Results}

\begin{table*}[htbp]
        \caption{Break-Even Analysis Summary for All Model-API Combinations (months) with Amortized Performance Differences}
        \label{tab:breakeven-summary}
        \centering
        \small
        \begin{tabular}{|l|c|c|c|c|c|c|}
            \hline
            \textbf{Open Model} & \textbf{GPT-5} & \textbf{Claude-4 Opus} & \textbf{Claude-4 Sonnet} & \textbf{Grok-4} & \textbf{Gemini 2.5 Pro} & \textbf{Range} \\
            \hline
            \multicolumn{7}{|c|}{\textit{Large Open Models}} \\
            \hline
            Kimi-K2 & 69.3 (-6.75\%) & 8.7 (\textbf{+1.83\%}) & 44.0 (+5.35\%) & 44.0 (-10.88\%) & 63.1 (-8.93\%) & 8.7-69.3 \\
            GLM-4.5 & 51.5 (-1.32\%) & 6.5 (+7.25\%) & 32.8 (+10.78\%) & 32.8 (-5.45\%) & 47.0 (-3.50\%) & 6.5-51.5 \\
            Qwen3-235B & 34.0 (\textbf{+0.45\%}) & 4.3 (+9.03\%) & 21.8 (+12.55\%) & 21.8 (\textbf{-3.68\%}) & 31.1 (\textbf{-1.73\%}) & 4.3-34.0 \\
            \hline
            \multicolumn{7}{|c|}{\textit{Medium Open Models}} \\
            \hline
            gpt-oss-120B & 30.9 (-5.47\%) & 3.9 (+4.20\%) & 19.8 (+8.67\%) & 19.8 (-11.10\%) & 28.2 (-9.27\%) & 3.9-30.9 \\
            GLM-4.5-Air & 34.0 (-4.75\%) & 4.3 (+3.83\%) & 21.8 (+7.35\%) & 21.8 (-8.88\%) & 31.1 (-6.93\%) & 4.3-34.0 \\
            Llama-3.3-70B & 17.8 (-27.88\%) & 2.3 (-19.30\%) & 11.4 (-15.78\%) & 11.4 (-32.00\%) & 16.2 (-30.05\%) & 2.3-17.8 \\
            \hline
            \multicolumn{7}{|c|}{\textit{Small Open Models}} \\
            \hline
            EXAONE 4.0 32B & 2.26 (-2.65\%) & 0.3 (+5.93\%) & 1.4 (+9.45\%) & 1.4 (-6.43\%) & 2.06 (-4.48\%) & 0.3-2.26 \\
            Qwen3-30B & 2.5 (-5.25\%) & 0.3 (+3.38\%) & 1.6 (+6.90\%) & 1.6 (-9.00\%) & 2.3 (-7.05\%) & 0.3-2.5 \\
            Magistral Small & 3.0 (-12.25\%) & 0.4 (-3.23\%) & 1.9 (\textbf{+0.28\%}) & 1.9 (-15.73\%) & 2.76 (-13.78\%) & 0.4-3.0 \\
            \hline
        \end{tabular}
    \end{table*}
    By substituting the values from Table~\ref{tab:benchmark} and Table~\ref{tab:gpu-comparison} into Equation~\eqref{equation:break-even}, we obtain the results summarized in Table~\ref{tab:breakeven-summary}, with cost details in Table~\ref{tab:hardware}. Our analysis across 54 deployment scenarios—covering nine open-source models and six commercial APIs—shows substantial variation in economic viability across model sizes.
    
    \textbf{Small Models.}
    Sub-30B models prove highly cost-effective, often reaching break-even within three months. Their low hardware cost enables smaller organizations to sustain competitive AI capabilities without ongoing subscription fees.
    
    \textbf{Medium Models.}
    Medium-scale deployments require moderate investment but achieve favorable returns over time. They offer a practical balance between performance and cost and are suitable for enterprises with steady, high-volume workloads.
    
    \textbf{Large Models.}
    Large-scale setups face the steepest economic barriers. Despite competitive accuracy, their high hardware and power demands extend break-even horizons. This challenge makes them viable only for organizations with sustained, large-scale inference needs.

    \subsection{Deployment Decision Framework}

    We analyzed break-even points by both model size and enterprise type. Small, medium, and large organizations have different needs for computing power, regulatory compliance, and financial resources. These factors shape whether on-premise deployment is practical. Table~\ref{tab:breakeven-summary} reports the quantitative results. The following sections explain what these results mean for organizations of different sizes.

    \subsubsection{Small Enterprises (SMEs)}
    
    For SMEs with limited budgets and moderate workloads ($<$10M tokens/month), small open-source models such as EXAONE 4.0 32B and Qwen3-30B offer the most viable entry point. Our results show that break-even can occur in as little as 0.3--3 months depending on the commercial baseline (e.g., Claude-4 Opus vs. GPT-5). This aligns with recent findings that SMEs prioritize cost savings and data control over absolute performance \cite{wu2025mia, overthink_cost}. The feasibility of deploying on consumer-grade GPUs (e.g., RTX 5090 at $\sim$\$2,000) further reduces capital barriers. Typical use cases include customer support automation, internal knowledge search, and lightweight document analysis—tasks where 30B-class models provide sufficient accuracy \cite{bai2023qwen, exaone2025}.
    
    \subsubsection{Medium Enterprises}
    
    Medium-scale enterprises (processing 10--50M tokens/month) represent the sweet spot for on-premise adoption. Medium models such as GLM-4.5-Air and Llama-3.3-70B demonstrate balanced economics, with break-even periods ranging from 3.8 to 34 months depending on provider comparison. Hardware requirements remain manageable (\$15k--\$30k for dual A100 setups), and throughput levels (120--220 tokens/sec) are sufficient for concurrent business workloads such as code assistance, analytics, and customer-facing applications. This tier benefits most from hybrid strategies, where sensitive workloads run locally and burst traffic leverages cloud APIs \cite{hao2024hybrid, chen2025}. Regulatory-driven industries such as healthcare and finance particularly favor medium-scale deployments for balancing compliance and cost \cite{desai2024}.
    
    \subsubsection{Large Enterprises}
    
    For large enterprises with extreme-scale workloads ($>$50M tokens/month), large open-source models (e.g., Qwen3-235B, Kimi-K2) become economically attractive, albeit with longer break-even horizons (3.5--69.3 months). While upfront investments exceed \$40k--\$190k, these organizations often already operate GPU clusters for other workloads, reducing incremental CapEx \cite{kim2025tco, inference_econ}. Use cases include enterprise-wide generative applications, advanced research, and domain-specific copilots requiring reasoning depth comparable to commercial APIs. However, when compared against aggressively priced providers such as Gemini 2.5 Pro, break-even can extend to 5--9 years, challenging the economic case unless privacy, sovereignty, or vendor lock-in are overriding concerns \cite{park2025survey, energy_considerations}. Thus, for this tier, non-financial factors (e.g., strategic autonomy, compliance) often weigh more heavily than pure cost.
    
\section{Conclusion}

This study evaluates 54 deployment scenarios to clarify the economic trade-offs of on-premise versus API-based LLM use. Results show that deployment economics are highly context-dependent and challenge common assumptions about local feasibility.

\begin{itemize}
    \item \textbf{Small-scale:} Break even within 3 months, making local deployment accessible for smaller organizations.  
    \item \textbf{Medium-scale:} Offer balanced performance and cost, with recovery typically within 6–24 months.  
    \item \textbf{Large-scale:} Require longer horizons (often beyond 2 years), feasible only for sustained, high-volume workloads.  
    \item \textbf{Pricing variability:} Differences among providers create large swings in cost-effectiveness, with premium tiers enabling faster local payback.
\end{itemize}

\textbf{Strategic Implications.}  
Deployment choices can be grouped into short-term (0–6 months), mid-term (6–24 months), and long-term ($\ge$24 months) investments, which offers organizations a practical lens to align cost recovery with strategic goals. The rapid evolution of models, hardware, and pricing means deployment is a continuous optimization problem rather than a one-time decision.

\textbf{Future Work.}  
Further research should empirically validate these break-even estimates and extend TCO models to include staffing and maintenance, and explore hybrid paradigms that combine economic efficiency with reliability. Continued benchmarking will be essential to track performance and cost convergence between open and commercial ecosystems.

Overall, this study situates break-even analysis within the broader landscape of evolving LLM technology and highlights both the growing accessibility of local deployment and the ongoing economic tension with cloud-based services.

\end{document}